\documentclass[twoside,leqno,twocolumn]{article}  
\usepackage{ltexpprt} 

\usepackage{cite}
\usepackage{amsmath}
\usepackage{amssymb}
\usepackage{varwidth}
\usepackage{graphicx}
\usepackage{epstopdf}
\usepackage{tabularx}
\usepackage{algorithmic}
\usepackage{algorithm}
\usepackage{wrapfig}

\begin{document}

\title{\Large Enforced Sparse Non-Negative Matrix Factorization}
\author{ Brendan Gavin\thanks{University of Massachusetts, Amherst} \footnotemark[2] \and Vijay Gadepally\thanks{MIT Lincoln Laboratory} \and  Jeremy Kepner\footnotemark[2] }
\maketitle

\begin{abstract} \small\baselineskip=9pt 
Non-negative matrix factorization (NMF) is a common method for generating topic models from text data. NMF is widely accepted for producing good results despite its relative simplicity of implementation and ease of computation. One challenge with applying NMF to large datasets is that intermediate matrix products often become dense, stressing the memory and compute elements of a system. In this article, we investigate a simple but powerful modification of a common NMF algorithm that enforces the generation of sparse intermediate and output matrices. This method enables the application of NMF to large datasets through improved memory and compute performance. Further, we demonstrate empirically that this method of enforcing sparsity in the NMF either preserves or improves both the accuracy of the resulting topic model and the convergence rate of the underlying algorithm.\end{abstract}

\section{Introduction}

\let\thefootnote\relax\footnotetext{This work is sponsored by the Assistant Secretary of Defense for Research and Engineering under Air Force Contract \#FA8721-05-C-0002. Opinions, recommendations, and conclusions are those of the authors and are not necessarily endorsed by the United States Government.}

A common analyst challenge is searching through large quantities of text documents to find interesting pieces of information. With limited resources, analysts often employ automated text-mining tools that highlight common terms or topics. The machine learning and natural language processing communities often refer to this as topic modeling. Topic modeling is a vast field in which there have been many fundamental contributions. For example, latent dirichlet allocation (LDA)~\cite{blei2003latent} uses Bayesian networks to model how a mixture of topics constitutes a document. Other common methods for topic modeling include the following: latent semantic analysis (LSA)~\cite{blei2003latent},  probabilistic latent semantic analysis (PLSA)~\cite{hofmann1999probabilistic}, and term frequency-inverse document frequency (TF-IDF)~\cite{gerard1983introduction}. More recently, non-negative matrix factorization is used as a technique for document classification and topic modeling.

\noindent \hrulefill 

\noindent \textbf{Definitions} \footnotesize

\begin{description}
\item[NMF:] Non-negative matrix factorization, i.e., $A=UV^T$
\item[A:] Term/document data matrix
\item[U:] Term/topic matrix, \ \ \textbf{V:} Document/topic matrix
\item[Topic:] A cluster of related documents or terms. Columns of $U$ are term topics and columns of $V$ are document topics.
\item[ALS:] Alternating least squares; an algorithm for finding the NMF.
\item[Residual:] Relative norm of the difference between $U$ matrices at subsequent iterations of ALS.
\item[Error:] Relative norm of the difference between data matrix $A$ and its NMF approximation $UV^T$
\item[NNZ:] Number of nonzeros
\end{description}
\hrulefill \normalsize

The non-negative matrix factorization (NMF) \cite{leeseungnature,paatero1994positive,lee2001algorithms,wang2013nonnegative} is a method of finding a latent variable model of non-negative data for the purpose of dimensionality reduction. It is frequently used to generate topic models of text data, which can be used for clustering related documents and terms from a collection of documents. While there are many methods for document clustering and topic modeling, the NMF has evolved as a popular tool because of its algorithmic simplicity, implementation ease, and computational benefits. As described in~\cite{gadepally2015gabb}, the NMF can be ideal for systems that are designed to perform fast matrix operations in order to support specifications for graph algorithms such as the GraphBLAS~\cite{kepner2014gabb, graphblas}. 

With the NMF, we have a data matrix $A$ whose entries are all greater than or equal to zero, and we seek to factorize it into a product of two matrices $U$ and $V$ whose entries are all non-negative. Mathematically, this property can be expressed as
\begin{align}
A\approx & \ UV^T\\
A\in \mathbb{R}^{n\times m},\ U&\in\mathbb{R}^{n\times k},\ V\in \mathbb{R}^{m\times k} \nonumber \\
\nonumber A\geq 0,\ U&\geq 0,\ V\geq 0
\end{align}
The rank $k$ of the decomposition is usually chosen to be much smaller than the rank of the original data matrix. In this way, much of the information in the original data set is discarded, with the intention that what remains will capture the most general, abstract relationships from the original data.

We find the NMF by solving a minimization problem in order to make the factorization $UV^T$ as close to the original data matrix as possible:
\begin{align}
\label{eqn_min_problem}
\min_{U,V}||A-U&V^T||, \ \ \ \text{s.t. } U\geq 0,\ V\geq 0 
\end{align}
There are a variety of ways to approach this problem as described in~\cite{berry2007algorithms}. Perhaps the most common method is to use the multiplicative update rules of Lee and Seung~\cite{lee2001algorithms}. The benefit of these update rules is that they are simple to implement and that analytical results can be established about the convergence properties of updates. Other methods include gradient descent algorithms (of which the multiplicative update rules of~\cite{lee2001algorithms} are an example) and the alternating non-negativitity constrained least squares (ANLS) method. In ANLS, alternating least squares problems are solved by using optimization methods to enforce the non-negativity constraint~\cite{kim2008nonnegative} in Equation (\ref{eqn_min_problem}). Many of these algorithms have analytical results regarding their convergence properties and have proven to be effective in practice; however, they tend to be slow to converge.

In the method we describe in this article, we have chosen to solve Equation (\ref{eqn_min_problem}) by using the conventional alternating least squares (ALS) algorithm combined with a projection step, as described in~\cite{berry2007algorithms}. In ALS, we hold one of the matrices $U$ or $V$ constant, and then solve for the other by using linear least squares. By repeating this process many times, alternating back and forth between solving for $U$ and solving for $V$, we hope to converge to a good approximation for the non-negative matrix factorization. The projection step enforces the non-negativity constraint of Equation (\ref{eqn_min_problem}) by setting all the negative entries of $U$ and $V$ to zero at each iteration, rather than by using constrained optimization methods as ANLS does. Because this is a projection onto the space of non-negative solutions, it is sometimes called projected alternating least squares. This algorithm is described in Algorithm~\ref{alg_nmf}.

\begin{algorithm}
\caption{Projected Alternating Least Squares}\label{alg_nmf}
\textbf{Input:} data matrix $A\in \mathbb{R}^{n\times m}$, initial guess $U_0\in \mathbb{R}^{n \times k}$
\textbf{Output:} Approximation $UV^T\approx A$, $U\in \mathbb{R}^{n \times k},$\\$\ V \in \mathbb{R}^{m \times k}$

\textbf{START:} Set $U=U_0$

\begin{description}
\item[ Do until convergence:] \ 
\begin{description}
\item[1.] Find $V$ using $V=A^TU(U^TU)^{-1}$, and set negative entries of $V$ to zero.
\item[2.] Find $U$ using $U=AV(V^TV)^{-1}$, and set negative entries of $U$ to zero.
\end{description}
\item[ end do]
\end{description}

\textbf{Output }$U$, $V$

\textbf{END}
\end{algorithm}

Because of the ad hoc means by which this algorithm enforces non-negativity, there are currently no analytic results regarding its convergence properties. Nonetheless, in practice, the projected ALS algorithm produces consistently good results. 

Our ultimate goal is to use NMF to work with very large datasets, and our first priority in implementation is improved performance. We prefer to use projected ALS because it is the fastest of the available methods for finding the NMF, and it can be implemented by using only basic linear algebra operations (specifically matrix-vector multiplication).

Because our goal is to work with very large datasets, we would also like to use sparse matrix storage formats for storing $U$ and $V$. Using sparse matrices offers significant performance benefits in terms of memory usage and computation time. The challenge is that, although the original data matrix $A$ is always very sparse, Algorithm \ref{alg_nmf} may produce intermediate $U$ and $V$ matrices that are dense (see Figure \ref{fig_sparse_dense} for examples). 

\begin{figure}[h]\centering
\begin{varwidth}{0.5\textwidth}
\textbf{Reuters-21578}\\

\begin{tabular}{ c | c }
\multicolumn{1}{c}{Matrix}&\multicolumn{1}{c}{Sparsity} \\
  \hline
$A$ & 99.65\% \\
$U$ & 61.0\% \\
$V$ & 61.0\% \\
$UV^T$ & 4.15\% 
\end{tabular}
\end{varwidth} \ \ \ \ \ \  \ \ 
\begin{varwidth}{0.5\textwidth}
\textbf{Wikipedia}\\

\begin{tabular}{ c | c }
\multicolumn{1}{c}{Matrix}&\multicolumn{1}{c}{Sparsity} \\ \hline
$A$ & 99.6\% \\
$U$ & 45.0\% \\
$V$ & 41.0\% \\
$UV^T$ & 11.0\%
\end{tabular}
\end{varwidth}
\caption{Tables showing the change in sparsity from the original data matrix $A$ to the NMF approximation of it, $UV^T$, using two different data sources. Sparsity is measured as the fraction of a matrix's entries that are exactly equal to zero. 
}\label{fig_sparse_dense}
\end{figure}

In the following section, we describe our approach for modifying Algorithm~\ref{alg_nmf} so that we can produce matrices that take advantage of the benefits of sparse matrix storage. In Section~\ref{sec_results}, we apply the resulting enforced sparsity NMF algorithm to several example datasets. We further show that the modified algorithm converges at least as well as Algorithm \ref{alg_nmf} and that it produces NMF topics that are empirically and qualitatively as accurate as those produced by the unmodified ALS algorithm. We also demonstrate a drawback of producing NMF matrices that are extremely sparse, and in Section 
\ref{sec_seqnmf} we discuss a couple of methods that can be used to alleviate that problem. We conclude in Section~\ref{conc}.

\section{Maintaining Sparsity in NMF}

Many studies have looked at producing sparse matrices through the NMF \cite{eggert2004sparse,kim2008nonnegative,hoyer2004non,peharz2012sparse,hoyer2002non}. Some studies use NMF to produce sparse matrices by adding sparsity constraints to the minimization problem (Equation~\ref{eqn_min_problem}), and others do so by adding terms to the minimization problem that penalize having larger numbers of nonzeros in the factorization. 
Some popular methods for ensuring sparsity include using Hoyer's sparsity measure \cite{hoyer2004non} or the $L_1$ norm of $U$ or $V$ \cite{kim2008nonnegative} as either constraints or penalty terms in the minimization problem (Equation~\ref{eqn_min_problem}). These approaches work well, but they require the use of algorithms for NMF that are slower than the projected ALS algorithm. 

Because we would like to factorize matrices that are derived from very large datasets, we prefer to use a method of producing sparse matrices that prioritizes performance
 over other concerns, such as provable optimality. 
To produce sparse intermediate and final matrices, we apply the same logic to sparsity that projected ALS applies to non-negativity: at each iteration of ALS, in order to ensure that either $U$ or $V$ has exactly $t$ nonzero entries in it, we set all the entries in that matrix to zero except for the $t$ largest ones. The resulting modification of Algorithm \ref{alg_nmf} is shown in Algorithm \ref{alg_alsnmf}.

\begin{algorithm}
\caption{Enforced Sparsity ALS}\label{alg_alsnmf}
\textbf{Input:} matrix $A\in \mathbb{R}^{n\times m}$, initial guess $U_0\in \mathbb{R}^{n \times k}$, maximum NNZ(U) $t_u$ and maximum NNZ(V) $t_v$

\textbf{Output:} Approximation $UV^T\approx A$, $U\in \mathbb{R}^{n \times k},$\\ $\ V \in \mathbb{R}^{m \times k}$

\textbf{START:} Set $U=U_0$

\begin{description}
\item[ Do until convergence:] \ 
\begin{description}
\item[1.] Find $V$ using $V=A^TU(U^TU)^{-1}$, and set negative entries of $V$ to zero.
\item[2.] Sort nonzero entries of $V$, keep only $t_v$ largest nonzeros in $V$
\item[3.] Find $U$ using $U=AV(V^TV)^{-1}$, and set negative entries of $U$ to zero.
\item[4.] Sort nonzero entries of $U$, keep only $t_u$ largest nonzeros in $U$.
\end{description}
\item[ end do]
\end{description}

\textbf{Output }$U$, $V$

\textbf{END}
\end{algorithm}

Herein, this algorithm is referred to as the enforced sparsity ALS, because we are producing sparse matrices by enforcing sparsity at each iteration of the ALS algorithm. In this method, we keep the $t$ largest entries of a given matrix by finding the magnitude of the $t^{\text{th}}$ largest entry and then setting all the entries with magnitudes lower than that of the $t^{\text{th}}$ largest entry to zero. This method requires slightly more computation than a simpler method of enforcing sparsity, which is to set all entries of a matrix that fall below an arbitrary threshold to zero; however, it has the benefit of allowing us to consistently set exactly the amount of sparsity that we want, regardless of the scaling of the matrices that we are working with.

\section{Enforced Sparsity NMF Results}
\label{sec_results}
We illustrate the results of using Algorithm \ref{alg_alsnmf} by applying it to data matrices that are derived from several different real datasets of text documents. For each of these datasets, we produce a term-document data matrix $A$, where each column represents a single document, each row represents a single term, and each entry $a_{ij}$ is the number of times that term $i$ occurs in document $j$. We discard terms from each document by using a stop word list, and we also discard any terms that appear only once in a particular dataset. We divide each row of the data matrix by the number of nonzero entries in that row in order to prevent our results from being biased by commonly used terms. All of the matrices that we use to produce our results -- including the $U$ and $V$ matrices -- are stored in the MATLAB sparse matrix storage format.

\subsection{Convergence Behavior}
\label{sec_convergence}
We examine the effect of sparsity enforcement on the convergence of ALS by examining the relative error and the relative residual of the NMF at each iteration of ALS.

\begin{figure}[h]
\begin{center}
\large
\textbf{NMF With and Without Sparsity Enforcement}\end{center}
\scriptsize
\begin{varwidth}{0.5\textwidth}
\includegraphics[width=\textwidth]{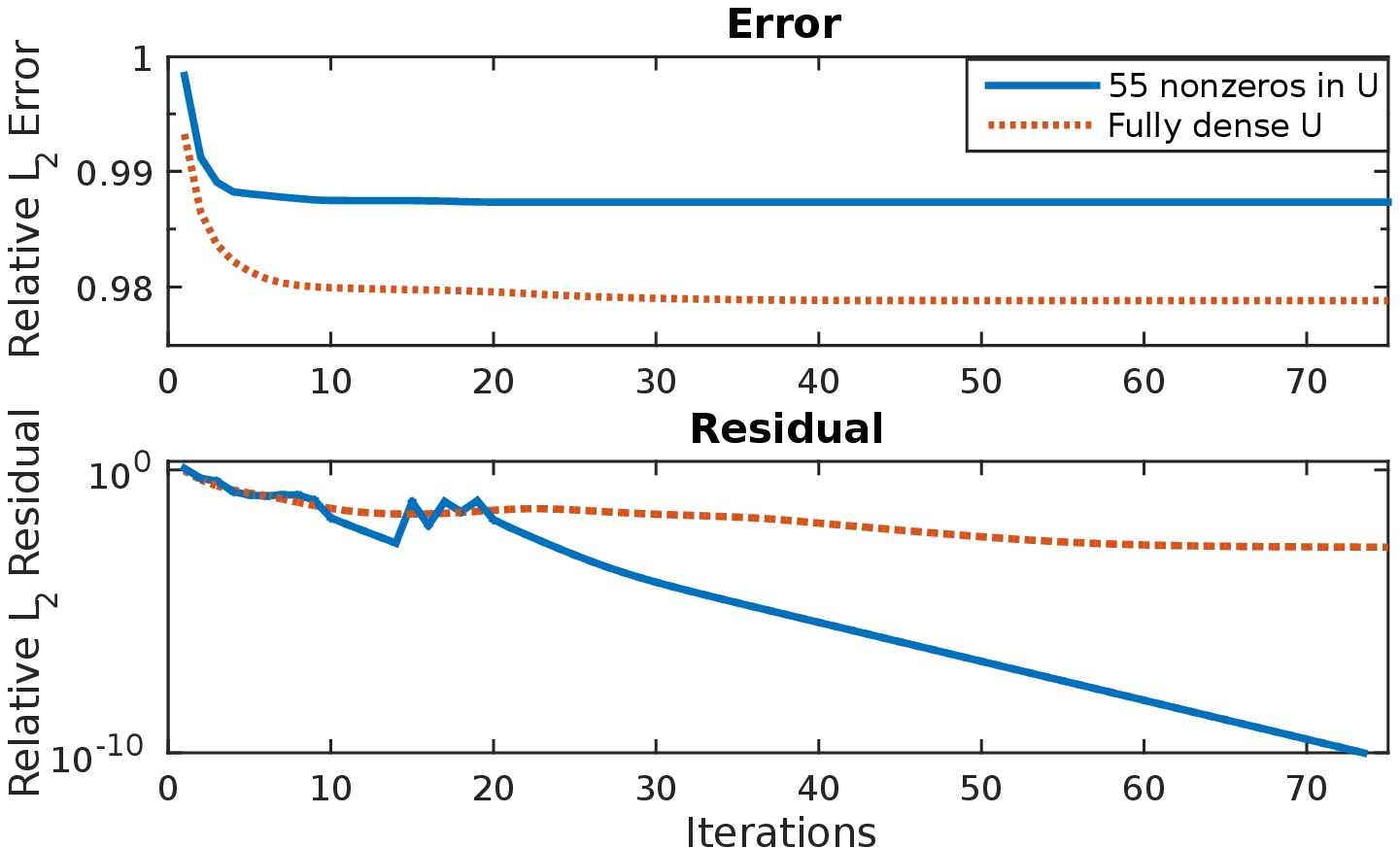}
\end{varwidth}
\begin{varwidth}{0.5\textwidth}
Sparsity Enforced $U$ Matrix (55 nonzeros for 5 topics):

\begin{tabular}{|c|c|c|c|c|}
\hline
Topic 1 & Topic 2 & Topic 3 & Topic 4 & Topic 5 \\ \hline \hline
miles & risk & coffee & repurchase & yen  \\ \hline
load & contracts & quotas & motors & firms  \\ \hline
factor & paper & ico & class & plaza  \\ \hline
revenue & proposals & crop & spending & currencies  \\ \hline
passenger & futures & colombia & buyback & movements  \\ \hline
\end{tabular}

\medskip

Fully Dense $U$ Matrix:

\begin{tabular}{|c|c|c|c|c|}
\hline
Topic 1 & Topic 2 & Topic 3 & Topic 4 & Topic 5 \\ \hline \hline
miles & paper & coffee & iran & senate\\ \hline
load & risk & crop & crude & baker\\ \hline
factor & proposals & quotas & opec & legislation\\ \hline
revenue & england & ico & iraq & vote\\ \hline
passenger & yen & producer & iranian & surplus\\ \hline
\end{tabular}
\end{varwidth}
\caption{Example NMF with and without sparsity enforcement. The plots on the top show the relative error and residual at each ALS iteration when using sparsity enforcement on the term/topic matrix $U$ and when allowing it to be fully dense. The tables on the bottom show the five terms with the largest magnitudes for each resulting topic.}\label{fig_reutex}
\end{figure}

Figure~\ref{fig_reutex} shows some of the results of using projected ALS, with and without sparsity enforcement, to find a five-topic NMF of a data matrix derived from the Reuters-21578 dataset (provided by~\cite{lewis}) using 1,985 documents with 6,424 different terms.
Figure \ref{fig_reutex} shows the relative residual and the relative approximation error at each iteration of Algorithm \ref{alg_alsnmf}. The relative residual $R=||U_{i}-U_{i-1}||/||U_{i}||$
is a measure of the difference between our current solution for $U$ and our solution for $U$ at the previous iteration. If $R$ reaches machine precision, then we know that we have reached a stable solution. We use the residual as a natural measure of convergence. The relative approximation error $E=||A-UV^T||/||A||$ is a measure of how close our NMF is to the original data matrix $A$ that reaches machine precision when the two are equal. We find that there is not necessarily a clear relationship between the approximation error and the quality of the topic model that our NMF produces, but it is a useful measure to see if an NMF algorithm is doing what it is supposed to be doing. If the algorithm is working, the error should decrease for a bit before reaching a steady value, the magnitude of which is determined by the rank of the factorization.

In Figure \ref{fig_reutex}, when the NMF is generated using the enforced sparsity ALS, the term/topic matrix $U$ is forced by Algorithm \ref{alg_alsnmf} to have only 55 nonzeros (in order to maintain sparsity); in the dense case, $U$ is allowed by Algorithm \ref{alg_nmf} to have any number of nonzeros. In this example, the run with a sparse $U$ converges more quickly than the fully dense version (as measured by the relative residual), and finishes with a higher relative $L_2$ error. Both runs qualitatively produce coherent topic terms, although the topics produced by each are somewhat different. 

The results described in Figure~\ref{fig_reutex} are representative of the empirical behavior of Algorithm \ref{alg_alsnmf}, in producing solutions that converge. They are also representative of the algorithm's behavior in the sense that Algorithm~\ref{alg_alsnmf} consistently produces results with higher numerical approximation error than does Algorithm~\ref{alg_nmf}. From our numerous tests, the higher numerical approximation error does not appear to produce poorer topic quality, as determined by examining the highest magnitude terms that belong to each topic.

\begin{figure}\center
\textbf{NMF with Sparse U, Sparse V, and Sparse U and V}

\includegraphics[width=\columnwidth]{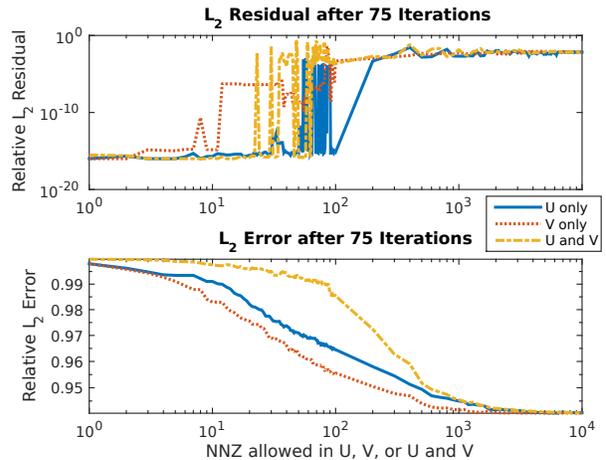}
\caption{Plots comparing the relative error and relative residual after 75 ALS iterations when enforcing sparsity for matrix $U$, for matrix $V$, and for both $U$ and $V$. 
}\label{fig_uvuv}
\end{figure}

The rate of convergence of Algorithm~\ref{alg_alsnmf} depends on how sparse we enforce each of the matrices $U$ and $V$ to be. Figure~\ref{fig_uvuv} shows the results of enforcing sparsity for only matrix $U$, for only matrix $V$, or for both matrices $U$ and $V$, for the same Reuters-21578 matrix. The vertical axes show the relative $L_2$ residual and relative $L_2$ error after 75 iterations of ALS, and the horizontal axes show the number of nonzeros allowed in the matrix that is forced to be sparse. 

The convergence behavior of enforced sparsity ALS falls broadly into two categories. For a low number of nonzeros (i.e., very sparse matrices), the algorithm tends to converge rapidly, and for a large number of nonzeros, it converges slowly, at the same pace as the non-modified conventional dense projected ALS. This result gives some context to the residual plot in Figure~\ref{fig_reutex}. Because limiting the $U$ matrix to have 55 nonzeros puts its convergence behavior roughly in between the rapid convergence region and the slow convergence region, the matrix factorization converges more rapidly than the dense case but not quickly enough to reach machine accuracy.

Figure \ref{fig_uvuv} seems to suggest that enforcing sparsity for just $U$, for just $V$, or for both $U$ and $V$, produces very similar results from the perspectives of error and convergence. However,

the distribution of nonzeros amongst the column vectors of $U$ and $V$ changes depending on which matrices are being made sparse. When we force one or both of $U$ and $V$ to be sparse (particularly when making them very sparse), the nonzeros in either matrix tend to end up unevenly distributed among the matrix's columns. As a result, some topics will have more terms or documents allocated to them, and other topics will have fewer. 

For example, allowing 50 nonzero entries in the term/topic matrix $U$ of a five-topic NMF typically does not result in having 10 nonzero entries in each column of that matrix. Instead, some columns will have very many nonzeros, and other columns will have very few. Table \ref{tab_wikiuneven} shows an example of the topic terms that are produced in this situation. The dataset used to produce this table is derived from the first 12,439 pages of the monthly Wikipedia database dump, with a total of 143,462 unique terms after stop words are filtered out. 

\begin{table}[H]\center

\textbf{Nonzeros Distributed Unevenly from Sparsity Enforcement}\\

\medskip

\footnotesize 
\begin{tabular}{|c|c|c|c|c|}
\hline
Topic 1 & Topic 2 & Topic 3 & Topic 4 & Topic 5 \\ \hline \hline
government & league & electrons & album & jewish\\ \hline
party &  & electron & band & jews\\ \hline
war &  & atoms & albums & judaism\\ \hline
elections &  & hydrogen &  & israel\\ \hline
president &  & isotopes &  & hebrew\\ \hline
\end{tabular}
\caption{Top five terms in each of five topics produced by NMF as applied to Wikipedia data, with term/topic matrix $U$ forced to have only 50 nonzeros. Nonzero entries are distributed unevenly amongst the columns of $U$ as a result.}\label{tab_wikiuneven}
\end{table}
The skew of the distribution of nonzeros in the column vectors of $U$ or $V$ is most severe when both the $U$ and the $V$ matrices are made very sparse. In this case, this skew may result in all of the nonzeros concentrated in a single column in each matrix. In Section~\ref{sec_seqnmf}, we discuss two methods for addressing this problem.


\subsection{Clustering Accuracy}
\label{clusteringaccurary}

An empirical measure of document clustering accuracy can be used to examine the effect of sparsity enforcement. To do so, we use a corpus of documents that consists of the abstracts of papers from five PubMed academic journals: BioMed Central (BMC) Bioinformatics, BMC Genetics, BMC Medical Education, BMC Neurology, and BMC Psychiatry. 
The resulting corpus, after stop words are removed, consists of 20,112 unique terms and 7,510 documents.

We can devise a measure of the accuracy of an NMF topic model for this dataset by considering each journal as defining an empirical topic, in the sense that an academic journal is, in fact, a cluster of related documents. We believe that it is reasonable to presume that a clustering algorithm, when applied to the abstracts of papers from academic journals, produces accurate clusters if it groups abstracts from the same journals (provided that the subject matter of each journal is sufficiently different from the subject matters of the others). Our choice of particular journals was made, in part, because they cover different and distinct topics.

We measure the accuracy of each topic by counting the number of pairs of documents that belong to a topic and that are from the same journal, and then by dividing that number by the total number of possible pairs of documents. For each topic in the NMF of this dataset, we consider a document as ``belonging'' to a topic if its corresponding entry in the $V$ matrix is nonzero. We consider a topic from the NMF to be perfectly accurate if all the documents that belong to that topic are from the same journal, in which case the number of same-journal document pairs is equal to the number of all possible pairs. On the other hand, a topic has the lowest possible accuracy if the documents that belong to it are uniformly distributed among the journals in the dataset. In that case, many of the pairs of documents belong to different journals, and the ratio of the number of same-journal pairs to the number of total possible pairs will be small.

We use the following expression for this measure of the accuracy of an NMF document topic:
\begin{equation}
\label{eqn_acc}
Acc=\frac{\sum_{i=1}^{n_D-1}\sum_{k=i+1}^{n_D}Jnl(i,k)}{\beta-\alpha}-\frac{\alpha}{\beta-\alpha}
\end{equation}
Where $\alpha$ and $\beta$ are given by
\begin{align}
\begin{split} 
&\alpha=\left \lfloor{n_D/n_J}\right \rfloor\left (\frac{n_J(\left \lfloor{n_D/n_J}\right \rfloor-1 )}{2}+n_D(\text{mod } n_J) \right )\\
&\beta=\frac{n_D(n_D-1)}{2}
\end{split}
\end{align}
The parameter $n_D$ is the number of documents that belong to the topic whose accuracy we are measuring (i.e., the number of nonzero entries in that column of the matrix $V$), $n_J$ is the number of journals that was used to make our dataset, and $Jnl(i,k)$ returns $1$ if document $i$ comes from the same journal as document $k$ and $0$ otherwise. The value of $\alpha$ is the number of same-journal pairs of documents when the documents belonging to a topic are uniformly distributed amongst all of the journals in the dataset; this will be the case when that topic's accuracy is the lowest. The value of $\beta$ is the maximum possible number of pairs of documents. 

The values $\alpha$ and $\beta$ scale and offset the measure (\ref{eqn_acc}) so that it is equal to 1 when all of the documents in a topic come from the same journal, and it is equal to 0 when they are perfectly uniformly distributed amongst the journals in the corpus. For cases in which a topic has only one or zero documents belonging to it, we set $Acc=1$. 

\begin{figure}\center
\textbf{Document Clustering Accuracy vs. NNZ}

\includegraphics[width=\columnwidth]{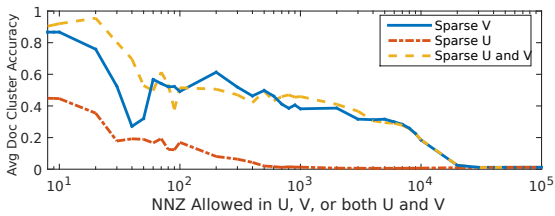}
\caption{Average document clustering accuracy versus number of nonzeros when using sparsity enforcement for finding the NMF of a data matrix derived from the abstracts of papers from PubMed journals. Accuracy for each individual topic is measured by using equation (\ref{eqn_acc}), and averaged over each of the topics in a 5 topic NMF. We show the results when enforcing sparsity for just the $U$ matrix, just the $V$ matrix, and both the $U$ and $V$ matrices.}\label{fig_acc_normal}
\end{figure}

Figure~\ref{fig_acc_normal} shows the results of applying the measure in Equation (\ref{eqn_acc}) to the NMF of the PubMed journal dataset. For these results, we perform 50 iterations of Algorithm~\ref{alg_alsnmf} in order to find a five-topic NMF (that is the largest number of correct topics that the NMF should find for five journals). As before, we calculate the NMF when enforcing various levels of sparsity for either the $U$ matrix, the $V$ matrix, or both the $U$ and $V$ matrices. 

We find that the accuracy is, in general, higher for sparser $U$ and $V$ matrices, with the accuracy being the lowest for the fully dense conventional NMF. This result is not surprising; we consider a document as ``belonging'' to a topic if it has any nonzero value in the corresponding entry of the $V$ matrix, and we don't take into account the fact that many of the entries for each document in a given topic have small magnitudes, indicating that they probably do not belong to that topic despite the fact that their value is nonzero. 

\begin{figure}\center
\textbf{Document Clustering Accuracy: Dense NMF vs. Enforced Sparsity NMF}

\includegraphics[width=\columnwidth]{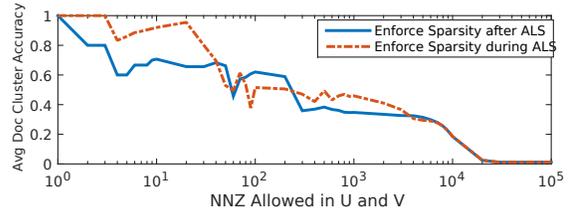}
\caption{Average document clustering accuracy, as measured by using Equation (\ref{eqn_acc}), versus the number of nonzeros in the $U$ and $V$ matrices. The plot curve labeled ``Enforce Sparsity during ALS'' shows the accuracy of the NMF produced by using Algorithm \ref{alg_alsnmf}, and the plot curve labeled ``Enforce Sparsity after ALS'' shows the accuracy of the NMF produced by using Algorithm \ref{alg_nmf}, where we have made the final NMF matrices sparse by enforcing sparsity only after the NMF has been calculated. By this measure, Algorithm \ref{alg_alsnmf} typically produces NMF document clusters that are at least as accurate as those produced by Algorithm \ref{alg_nmf}.}\label{fig_acc_compare}
\end{figure}

This measure is still useful, though, in allowing us to compare the accuracy of dense NMF to the accuracy of the proposed enforced sparse NMF. We can measure the accuracy of a topic from dense NMF by defining some threshold value below which we would consider the entries of $V$ to be zero, and then applying the accuracy measure in Equation (\ref{eqn_acc}) to the newly sparse $V$. To use Equation (\ref{eqn_acc}) with dense matrices, we enforce sparsity after completing ALS iterations when finding the NMF, instead of enforcing sparsity during each ALS iteration. Figure~\ref{fig_acc_compare} shows the result of measuring the document clustering accuracy for the NMF of the PubMed dataset when we enforce sparsity during each iteration of ALS, as is done in Algorithm~\ref{alg_alsnmf}, and when we enforce sparsity only after we have finished all our ALS iterations using Algorithm \ref{alg_nmf}.

The accuracy of the document clustering is approximately the same, regardless of whether we enforce sparsity during ALS or after finishing ALS. This result is encouraging because it suggests that we are producing equally good topics by enforcing sparsity at each iteration, but it raises an additional question: if we can make our final NMF just as sparse, and just as accurate, by enforcing sparsity only once after we have already calculated the NMF, why do it at each iteration? 

\begin{figure}[H]\center
\textbf{Maximum NNZ Stored When Calculating Sparse NMF}
\includegraphics[width=\columnwidth]{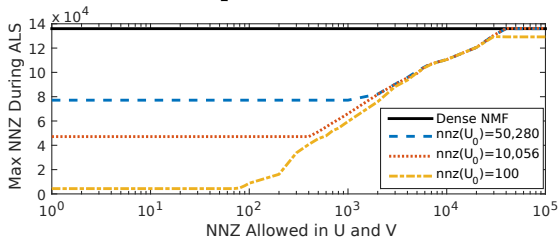}
\caption{Plot curves showing the maximum number of nonzeros that need to be stored for the $U$ and $V$ matrices combined when using Algorithm \ref{alg_alsnmf} to calculate the NMF of the PubMed dataset. Results are shown for several initial guesses with varying numbers of nonzeros. }\label{fig_maxnnz}
\end{figure}

The reason for enforcing sparsity at each iteration is that we would like to keep our matrices as sparse as possible at all times when computing the NMF, in order to reduce the memory footprint of both the final result matrices and the intermediate matrices that are generated during the iterations that we use to calculate NMF. By enforcing sparsity at each iteration, we can reduce our memory footprint considerably throughout the calculation. Exactly how much we are able to reduce the memory footprint throughout our calculation depends on the level of sparsity that we enforce, and the on the level of sparsity in our initial guess. Figure \ref{fig_maxnnz} shows the maxmimum number of nonzeros in $U$ and $V$ combined when we enforce sparsity for both matrices during the calculation of the NMF for the PubMed dataset. We show the results for various levels of sparsity in the initial guess $U_0$. Unsurprisingly, the maximum number of nonzeros that we need to store during our calculations is determined by the sparsity of the initial guess when we are forcing $U$ and $V$ to have very small numbers of nonzeros, and it is determined by the level of sparsity that we enforce in $U$ and $V$ when they are allowed to be more dense than the initial guess.

This example demonstrates the memory benefits of using sparsity enforcement at each iteration: we can reduce the amount of memory that we need to use to store $U$ and $V$ by more than an order of magnitude.
\section{Improving Clustering and Sequential ALS NMF}
\label{sec_seqnmf}

In Section~\ref{sec_convergence} we showed that, when using Algorithm \ref{alg_alsnmf}, we may produce an NMF that has unevenly distributed terms and documents among its topics. This problem is typically only as severe as it is in the results in Table \ref{tab_wikiuneven} when forcing our matrices to be very sparse but, in our opinion, the fact that Algorithm \ref{alg_alsnmf} can be used to produce matrices of such extreme sparsity is one of its benefits, and so we have considered two ways of alleviating this problem.

One particularly straightforward method to ensure even distribution of nonzeros among columns of our matrices is to enforce sparsity for each column individually rather than for the matrix as a whole. This procedure results in convergence and produces good topic models, but at the cost of reduced performance. This performance reduction occurs because, for most sparse matrix storage formats, addressing the entries of specific columns of a matrix is slower than addressing the entries of that matrix irrespective of which column they belong to. This additional time to address the contents results in reduced performance.

Another way to ensure an even distribution of nonzeros among the columns of our matrices is to find the NMF topics sequentially by converging one topic at a time. We can do this by considering the NMF using block matrices
\begin{equation}
A\approx UV^T=[U_1\ U_2][V_1\ V_2]^T=U_1V_1^T+U_2V_2^T
\end{equation}

\noindent where $U_1$ and $V_1$ are matrices whose column vectors consist of previously converged NMF topics, and $U_2$ and $V_2$ are single column vectors that represent the new topics that we seek to find. We can derive the update rules for finding $U_2$ and $V_2$ by rewriting the original minimization problem using the matrix blocks:
\begin{equation}
\min ||A-UV^T||^2_2=\min ||A-U_1V_1^T-U_2V_2^T||^2_2 
\end{equation}
The solution for one of $U_2$ or $V_2$, while holding the other constant, is then a modified least squares solution:
\begin{align}
\label{eqn_reg_seq_alsV}
&V_2=(A^TU_2-V_1U_1^TU_2)(U_2^TU_2)^{-1}\\
\label{eqn_reg_seq_alsU}
&U_2=(AV_2-U_1V_1^TV_2)(V_2^TV_2)^{-1}
\end{align}
We can find a $k$-topic NMF using these update rules by doing projected ALS $k$ times; we store each new topic that we generate as additional columns of the matrices $U_1$ and $V_1$ and then find the next topic by doing projected ALS again using Equations (\ref{eqn_reg_seq_alsV}) and (\ref{eqn_reg_seq_alsU}). We call this procedure sequential ALS because we are finding our NMF as a sequence of individual topics rather than as a block of topics. The full algorithm for sequential ALS is given in Algorithm \ref{alg_seq_nmf}.
\begin{algorithm}
\caption{Sequential ALS NMF}\label{alg_seq_nmf}
\textbf{Input:} data matrix $A\in \mathbb{R}^{n\times m}$, initial guess $U_0\in \mathbb{R}^{n \times k_2}$, maximum NNZ(U) $t_u$ and maximum NNZ(V) $t_v$, topics per block $k_2$, number of blocks $\eta$ (total number of topics $k$ = $\eta \times k_2$)

\textbf{Output:} Approximation $UV^T\approx A$, $U\in \mathbb{R}^{n \times k},$\\$\ V \in \mathbb{R}^{m \times k}$
\\
\textbf{START:} Find $U_1\in \mathbb{R}^{n \times k_2}$ and $V_1\in \mathbb{R}^{m \times k_2}$ such that $U_1V_1^T\approx A$ using Algorithm \ref{alg_alsnmf}, using $U_0$ as initial guess.
\begin{description}
\item[For] $i=2$ \textbf{to} $\eta$
\begin{description}
\item[]Set $U_2=U_0$
\item[ Do until convergence:] \ 
\begin{description}
\item[1.] Find $V_2$ using Equation (\ref{eqn_reg_seq_alsV}), and set negative entries of $V_2$ to zero.
\item[2.] Sort nonzero entries of $V_2$, keep only $t_v$ largest nonzeros in $V_2$
\item[3.] Find $U_2$ using Equation (\ref{eqn_reg_seq_alsU}), and set negative entries of $U_2$ to zero.
\item[4.] Sort nonzero entries of $U_2$, keep only $t_u$ largest nonzeros in $U_2$
\end{description}
\item[ end do]
\end{description}

Append the column vectors of $U_2$ and $V_2$ to the matrices $U_1$ and $V_1$ respectively, increasing their rank by $k_2$: $U_1=[U_1\ U_2]$, $V_1=[V_1\ V_2]$.

\item[end for]
\end{description}

\textbf{Output }$U=U_1$, $V=V_1$

\textbf{END}
\end{algorithm}

A similar procedure was previously proposed in~\cite{gupta2010additive}, where the author uses it as part of a method for generating a high-rank NMF in order to reconstruct missing data. The article in~\cite{gupta2010additive} does not use any means of ensuring sparsity in the resulting NMF, however, which would be necessary for the proposed algorithm to be used successfully on realistically large data sets.

\begin{figure}[h] \center
\textbf{Document Clustering Accuracy with Sequential and Column-wise Topic Sparsity}

\includegraphics[width=\columnwidth]{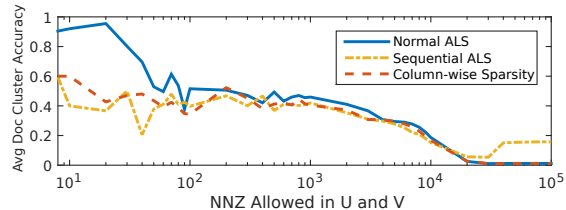}
\caption{Mean document clustering accuracy as measured by Equation (\ref{eqn_acc}) when calculating a five-topic NMF of the PubMed dataset by using either Algorithm \ref{alg_seq_nmf} or Algorithm \ref{alg_alsnmf} with sparsity enforced for each column of $U$ and $V$ individually. }\label{fig_docAccMethods}
\end{figure}

Figure \ref{fig_wikieven} shows the results of enforcing sparsity column by column and the results of using sequential NMF for the same Wikipedia data matrix that was used to produce Table~\ref{tab_wikiuneven}. Both methods produce an even distribution of nonzero entries among the columns of our matrices. Column-wise sparsity enforcement yields good topic terms, and sequential ALS yields good topic terms with the exception of topic 4. This behavior is consistent with what we have observed for the sequential ALS algorithm. While the algorithm often produces good topic terms, it is less robust than the typical projected ALS is. Figure~\ref{fig_docAccMethods} shows the accuracy of the NMF document topics when we use sequential ALS and column by column sparsity enforcement on the PubMed dataset, as measured by using Equation (\ref{eqn_acc}). By this measure, both methods produce document clusters that are approximately as accurate as the document clusters produced when we use the unmodified Algorithm~\ref{alg_alsnmf}.

\begin{figure} \center
\textbf{Sparsity Enforcement with Even Nonzero Distribution}
\scriptsize
\medskip

Enforce Sparsity by Column
\begin{tabular}{|c|c|c|c|c|}
\hline
Topic 1 & Topic 2 & Topic 3 & Topic 4 & Topic 5 \\ \hline \hline
government & proteins & electrons & album & jewish\\ \hline
party & protein & electron & band & jews\\ \hline
war & cells & atoms & music & judaism\\ \hline
president & cell & atom & albums & hebrew\\ \hline
election & dna & hydrogen & songs & torah\\ \hline
\end{tabular}

\medskip

Enforce Sparsity with Sequential ALS
\begin{tabular}{|c|c|c|c|c|}
\hline
Topic 1 & Topic 2 & Topic 3 & Topic 4 & Topic 5 \\ \hline \hline
government & city & album & film & game \\ \hline
war & population & band & church & games \\ \hline
party & airport & music & empire & players \\ \hline
military & census & albums & country & team \\ \hline
soviet & county & songs & united & league \\ \hline
\end{tabular}
\caption{Top five terms for each of five topics from Wikipedia data. This time we enforce sparsity for each column individually or by using sequential topic generation, with a limit of 10 nonzero entries per topic (for a total of 50 nonzero entries in the $U$ matrix). Compare with Table \ref{tab_wikiuneven}; here our topic terms are evenly distributed. 
}\label{fig_wikieven}
\end{figure}

\begin{figure} \center
\textbf{Time Scaling for Sparsity Enforcement}

\includegraphics[width=\columnwidth]{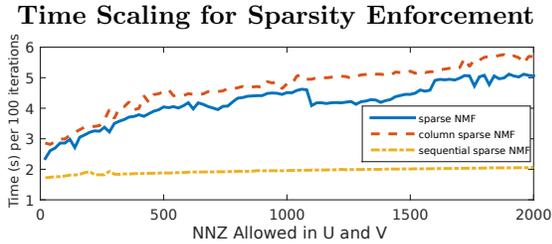}
\caption{Time required for 100 ALS iterations when finding a 5 topic NMF for the PubMed dataset. Results are shown using sparsity enforcement for the whole $U$ and $V$ matrices at once, for each column of $U$ and $V$ individually, and for columns of $U$ and $V$ generated sequentially. For the normal and column-wise NMF results, 100 projected ALS iterations are performed. For the Sequential ALS NMF, 20 iterations are performed for each of 5 topics that are generated, for a total of 100 ALS iterations.}\label{tab_sparsePerform}
\end{figure}

In order to compare the performance of sequential NMF and column-wise sparsity enforcement, Figure~\ref{tab_sparsePerform} shows the time required for 100 ALS iterations for finding a five-topic NMF for the PubMed journal dataset, using normal projected ALS NMF with sparsity enforcement, projected ALS NMF with column-wise sparsity enforcement, and sequential ALS. Enforcing sparsity for each column individually takes longer than doing so for the entire matrix at once, as we would expect when using sparse matrix formats. Sequential ALS is quite a bit faster than the other two methods. This result is not surprising, as sequential ALS does not require the use of a matrix inverse when $U_2$ and $V_2$ of Equations (\ref{eqn_reg_seq_alsV}) and (\ref{eqn_reg_seq_alsU}) have rank 1, as they do here; in that case, the inverse amounts to floating point division. Despite indications that sequential ALS provides less coherent term topics than the regular ALS NMF, this improvement in runtime suggests that it bears further investigation.

\section{Conclusion}
\label{conc}

In this article, we described a method to enforce sparsity in the computation of the NMF of a potentially large dataset. By setting the level of sparsity in our matrices that we desire at each iteration of alternating least squares, we can substantially reduce the memory resources that are required to compute NMF without sacrificing additional compute resources. In spite of the simplicity of the proposed approach, our experiments indicate that the algorithm converges at least as quickly and reliably as regular, dense projected ALS and that the NMF topic models that it produces are similarly accurate. 

Using this method, we can produce an NMF with matrices of fairly extreme sparsity, relatively inexpensively. In doing so we find that we produce NMF topic models with very unevenly distributed terms and documents among the topic clusters. We can produce topic models with perfectly evenly distributed topic terms and documents by enforcing sparsity for each topic individually. If we do this directly by enforcing sparsity for each column of $U$ and $V$ individually, then we find that we produce good term topics at the cost of sacrificing performance resulting from the slowdown caused by accessing sparse matrix formats column by column. If we do this by converging each topic sequentially, we produce quality term topics less reliably but with a considerable improvement in speed. Improving these techniques so that we do not have to sacrifice computation time or topic quality is a subject of continuing research.

\bibliographystyle{siam}

\bibliography{nmfPaper_siamDM}
\end{document}